\pdfoutput=1
\documentclass[11pt]{article}
\usepackage{arabtex}
\usepackage{utf8}
\usepackage{booktabs}
\usepackage{multirow}
\usepackage{array}
\usepackage{amsmath}
\usepackage{xcolor}
\usepackage{amssymb}
\usepackage{makecell}
\usepackage{pifont}
\usepackage{graphicx} 
\usepackage{lipsum}   
\usepackage{comment}
\usepackage{xurl}

\usepackage[final]{acl_natbib}
\usepackage{times}
\usepackage{latexsym}
\usepackage[T1]{fontenc}

\usepackage[utf8]{inputenc}

\usepackage{microtype}

\usepackage{inconsolata}

\usepackage{graphicx}
\definecolor{darkgreen}{rgb}{0.0, 0.5, 0.0} 
\definecolor{verylightgray}{rgb}{0.97, 0.97, 0.97}
\newcommand{\cmark}{\textcolor{darkgreen}{\scalebox{1}[1.0]{\ding{51}}}}
\newcommand{\xmark}{\textcolor{red}{\ding{55}}}  

\title{
    \raisebox{-0.3cm}{\includegraphics[width=1.2cm]{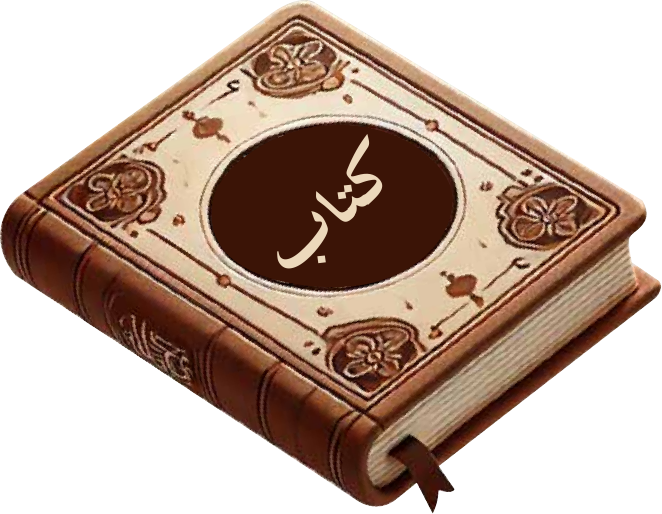}}  
    \hspace{0.01cm} \textbf{\Large KITAB-Bench: A Comprehensive Multi-Domain Benchmark for Arabic OCR and Document Understanding}
}

\author{ 
 Ahmed Heakl$^{\spadesuit \clubsuit}\footnotemark[1]$\hspace{0.5mm},
 Abdullah Sohail$^{\spadesuit}\footnotemark[1]$\hspace{0.5mm},
 Mukul Ranjan$^{\spadesuit}\footnotemark[1]$\hspace{0.5mm}, 
 Rania Hossam$^{\spadesuit}\footnotemark[1]$\hspace{0.5mm} \\
 \bf{Ghazi Ahmad$^\spadesuit$\hspace{0.5mm}, 
  Mohamed El-Geish$^{\clubsuit}$\hspace{0.5mm},
 Omar Maher$^{\clubsuit}$\hspace{0.5mm},  
 Zhiqiang Shen$^{\spadesuit}$}\hspace{0.5mm} \\
 \bf{Fahad Khan$^{\spadesuit \heartsuit}$\hspace{0.5mm},
 Salman Khan$^{\spadesuit \diamondsuit}$}\hspace{0.2mm}\\
$^\spadesuit$ MBZUAI \hspace{1.0mm}
$^\clubsuit$Monta AI \hspace{1.0mm}
$^\heartsuit$Linköping University\hspace{1.0mm}
$^\diamondsuit$Australian National University\\
 \quad\texttt{\{ahmed.heakl,mabdullah.sohail,mukul.ranjan,salman.khan\}@mbzuai.ac.ae} \\
 \quad\texttt{\{geish,omar\}@monta.ai} \\
 \url{https://mbzuai-oryx.github.io/KITAB-Bench/}
}

\begin{document}
\maketitle
\renewcommand{\thefootnote}{\fnsymbol{footnote}}
\footnotetext[1]{\ Equal Contributions}

\begin{abstract}
With the growing adoption of Retrieval-Augmented Generation (RAG) in document processing, robust text recognition has become increasingly critical for knowledge extraction. 
While OCR (Optical Character Recognition) for English and other languages benefits from large datasets and well-established benchmarks, Arabic OCR faces unique challenges due to its cursive script, right-to-left text flow, and complex typographic and calligraphic features. 
We present \textbf{\texttt{KITAB-Bench}}, a comprehensive Arabic OCR benchmark that fills the gaps in current evaluation systems.
Our benchmark comprises 8,809 samples across 9 major domains and 36 subdomains, encompassing diverse document types including handwritten text, structured tables, and specialized coverage of 21 chart types for business intelligence.
Our findings show that modern vision language models (such as GPT-4o, Gemini, and Qwen) outperform traditional OCR approaches (such as EasyOCR, PaddleOCR, and Surya) by an average of $60\%$ in the character error rate (CER).
Furthermore, we highlight significant limitations of current Arabic OCR models, particularly in PDF-to-Markdown conversion, where the best model Gemini-2.0-Flash achieves only $65\%$ accuracy. This underscores the challenges of accurately recognizing Arabic text, including issues with complex fonts, numeral recognition errors, word elongation, and table structure detection.
This work establishes a rigorous evaluation framework that can drive improvements in Arabic document analysis methods and bridge the performance gap with English OCR technologies.
\end{abstract}

\section{Introduction}
\begin{figure}[t]
    \centering
    \includegraphics[width=0.40\textwidth]{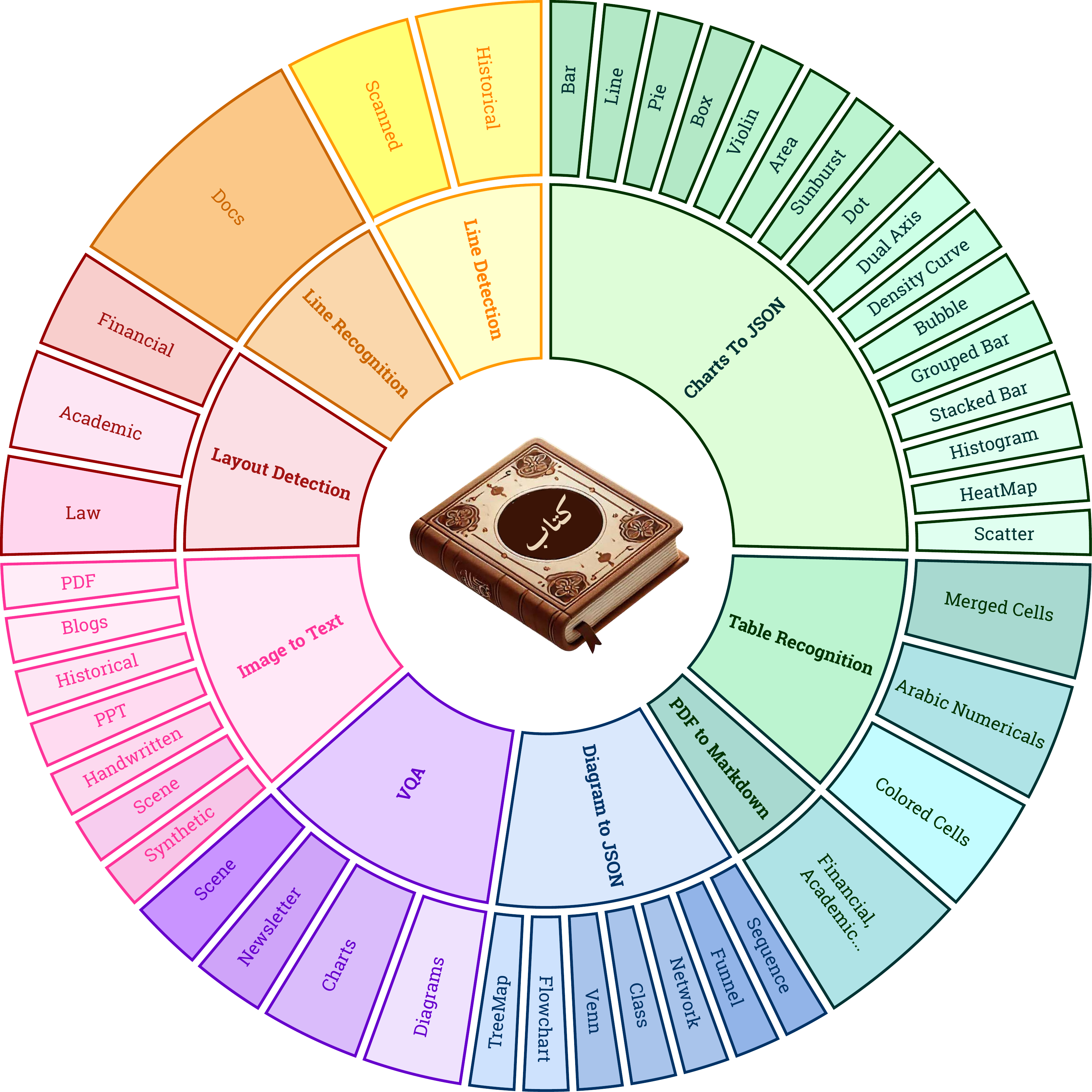}
    \caption{Overview of the core domains and sub-domains in KITAB-Bench. Our benchmark spans nine major domains (e.g., OCR, charts to JSON, table recognition) and 36 sub-domains (e.g., scanned text, handwritten text, various chart types), providing a comprehensive evaluation framework for modern Arabic document processing and analysis.}
    \label{fig:task_taxonomy}
\end{figure}

With the upsurge in adoption of Retrieval-Augmented Generation (RAG) based systems for document processing, the quality of document ingestion pipelines has become increasingly critical. Optical Character Recognition (OCR) plays a crucial role in this pipeline, enabling the conversion of physical documents into machine-readable text and databases for enabling effective knowledge retrieval. Although significant progress has been made in the multilingual OCR \cite{easyocr, fu2024ocrbench, wei2024general, smith2007overviewtesseract}, with comprehensive datasets like PubLayNet \cite{zhong2019publaynet}, DocBank \cite{li2020docbank}, M6Doc \cite{cheng2023m6doc}, and DocLayNet \cite{doclaynet2022}, Arabic OCR continues to lag behind. This gap is largely due to the unique challenges of the Arabic script, including its cursive nature, complex typography, and right-to-left text orientation.
\begin{figure*}[t]
    \centering
    \includegraphics[width=0.99\textwidth]{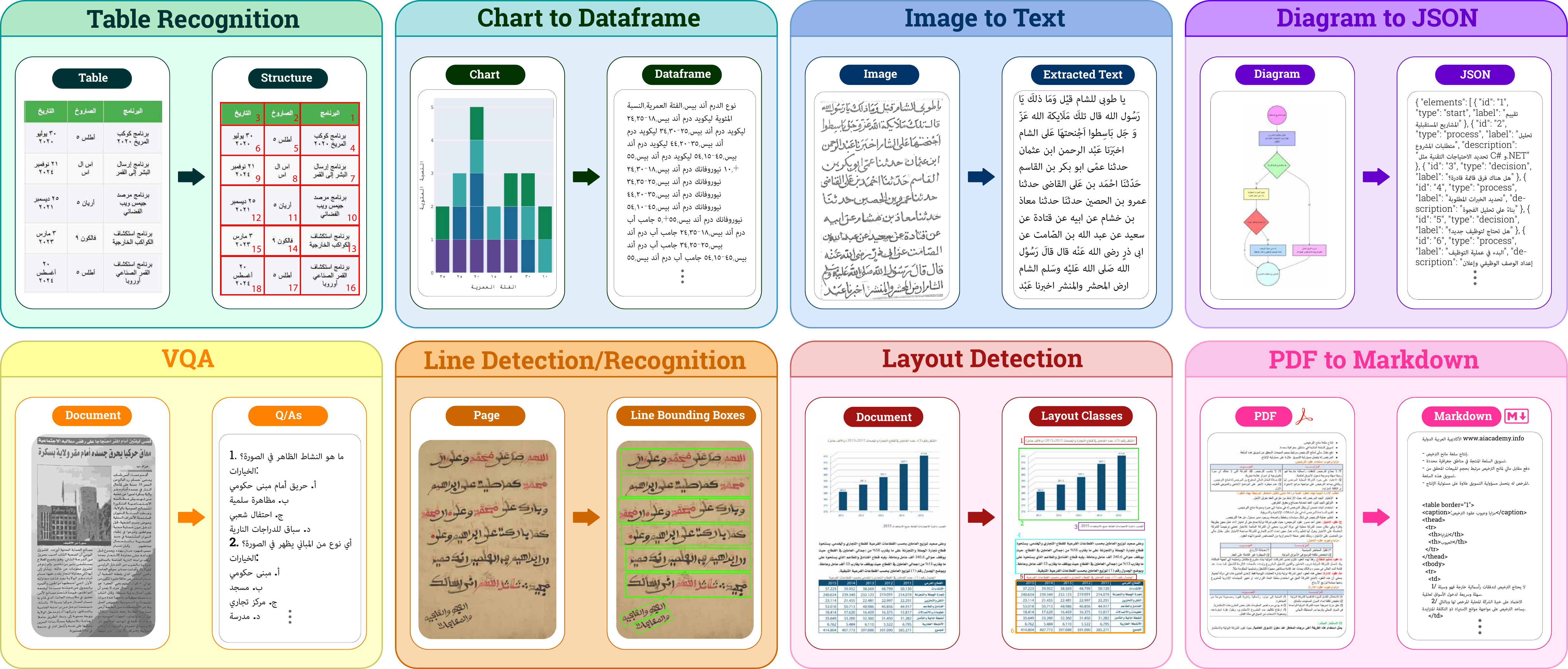}
    \caption{Overview of different tasks in our benchmark: Eight key components illustrating the task inputs and outputs for table recognition, chart understanding, text recognition, diagram analysis, VQA, line detection, layout analysis, and PDF-to-Markdown conversion, complete with input/output examples for each task.}
    \label{fig:pipeline_task_overview}
\end{figure*}

\begin{table}[htbp]
\centering
\resizebox{\linewidth}{!}{
\begin{tabular}{lccccc}
\hline
\textbf{Domain/} & \textbf{EXAMS-V$^*$} & \textbf{Camel-} & \textbf{MIDAD$^\dagger$} & \textbf{KHATT} & \textbf{KITAB-} \\
 \textbf{Characteristics} &  & \textbf{Bench} & \textbf{} & \textbf{} & \textbf{Bench (Ours)} \\
\hline
PDF to Markdown & \xmark & \xmark & \xmark & \xmark & \cmark \\
Layout Detection & \xmark & \xmark & \xmark & \xmark & \cmark \\
Line Detection & \xmark & \xmark & \xmark & \xmark & \cmark \\
Line Recognition & \xmark & \cmark & \xmark & \xmark & \cmark \\
Table Recognition & \xmark & \xmark & \xmark & \xmark & \cmark \\
Image to Text & \cmark & \cmark & \cmark & \cmark & \cmark \\
Charts to JSON & \xmark & \xmark & \xmark & \xmark & \cmark \\
Diagram to Code & \xmark & \xmark & \xmark & \xmark & \cmark \\
VQA & \cmark & \cmark & \xmark & \xmark & \cmark \\

Handwritten Samples & \xmark & \xmark & \cmark & \cmark & \cmark \\

Open Source & \cmark & \cmark & \xmark & \cmark & \cmark \\
\hline
Total Samples (\#) & 823 & 3,004 & 29,435 & 5,000 & 8,809 \\
\hline
\end{tabular}
}

\caption{Comparison of Arabic OCR Benchmarks Across Different Domains. Benchmarks compared: LaraBench \cite{abdelali2023larabench}, CamelBench \cite{ghaboura2024camel}, MIDAD \cite{bhatia2024qalam}, KHATT \cite{mahmoud2014khatt}, and KITAB-Bench (Ours).  ($*$: Only the Arabic  samples are considered.) ($\dagger$: The test set of the dataset is considered.)}
\label{tab:ocr_domain_comparison}
\vspace{-5pt}
\end{table}

\noindent Existing Arabic OCR datasets (Table~\ref{tab:ocr_domain_comparison}), like KHATT~\cite{mahmoud2014khatt} and IFN/ENIT~\cite{pechwitz2002ifn} focus mainly on handwritten text, whereas APTI~\cite{slimane2009new} covers only specific aspects of printed text. These efforts fail to address advanced document processing challenges such as table parsing, font detection, and numeral recognition. Arabic benchmarks like CAMEL-Bench~\cite{ghaboura2024camel} and LAraBench~\cite{abdelali2023larabench} evaluate large multimodal and language models, but they give limited attention to document understanding tasks. Consequently, there remains a need for a more comprehensive framework to systematically evaluate and compare Arabic OCR solutions. Our benchmark addresses these gaps by offering diverse document types and evaluation tasks to facilitate in-depth assessments of modern OCR systems.

We present KITAB-Bench, a comprehensive Arabic OCR benchmark spanning 9 domains and 36 sub-domains. Our framework evaluates layout detection (text blocks, tables, figures), multi-format recognition (printed/handwritten text, charts, diagrams), and structured output generation (HTML tables, DataFrame charts, markdown). This enables rigorous assessment of both basic OCR capabilities and advanced document understanding tasks.


The contributions of this work include (1) A comprehensive Arabic OCR benchmark covering multiple document types and recognition tasks.
(2) Detailed evaluation metrics for assessing performance across different document understanding challenges. We also propose CharTeX and CODM metric to evaluate chart extraction and diagram extraction respectively.
(3) Baseline results for popular OCR systems and Vision Language Models (VLMs), highlighting current limitations and areas for improvement.
(4) A standardized framework for comparing Arabic OCR systems, facilitating future research and development.

\begin{figure*}[t]
    \centering
    \includegraphics[width=0.97\textwidth]{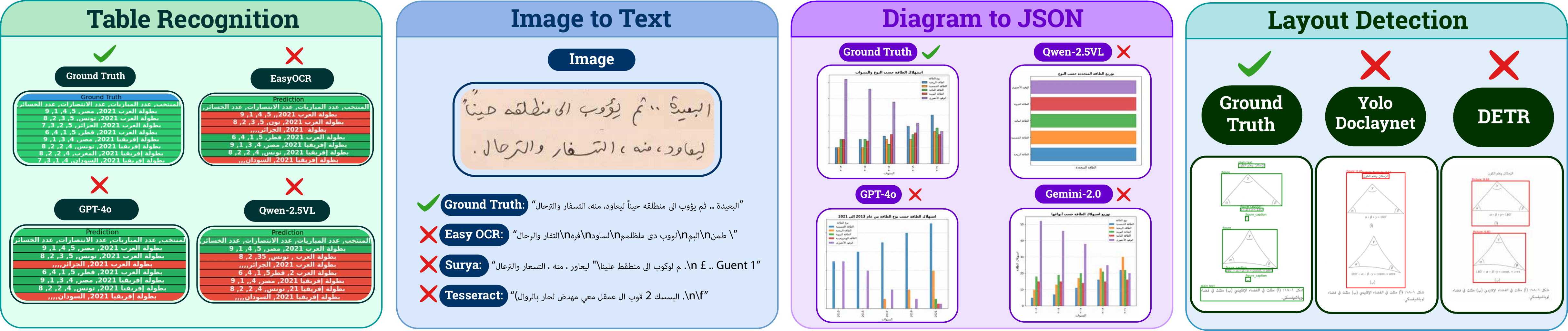}
    \caption{Comparison of model performance across four document understanding tasks (Table Recognition, Image to Text, Diagram to JSON, and Layout Detection) showing successful and failed cases for different models including Ground Truth, EasyOCR, GPT-4, Qwen, Surya, Tesseract, Yolo, and DETR on Arabic document benchmark data.}
    \label{fig:model_comparison}
\end{figure*}

\section{Related Work}
The development of robust Optical Character Recognition (OCR) systems has been extensively studied across document layout analysis \cite{zhao2024doclayout, shen2021layoutparser, paruchuri2024surya, easyocr, auer2024docling, li2020docbank}, table detection \cite{li2019tablebank, paliwal2019tablenet, nassar2022tableformer, li2021gfte, schreiber2017deepdesrt}, and document understanding \cite{staar2018corpus, weber2023wordscape, livathinos2021robust}. While English OCR benefits from rich datasets like PubLayNet \cite{zhong2019publaynet}, DocBank \cite{li2020docbank}, M6Doc \cite{cheng2023m6doc}, and DocLayNet \cite{doclaynet2022}, Arabic lacks standardized benchmarks for diverse fonts and layouts. Recent efforts like MIDAD \cite{bhatia2024qalam} curates extensive training data for Arabic OCR and handwriting recognition, while Peacock \cite{alwajih2024peacock} introduces culturally-aware Arabic multimodal models. Existing resources such as CAMEL-Bench \cite{ghaboura2024camel}, LAraBench \cite{abdelali2023larabench}, MADAR \cite{bouamor2018madar}, OSACT \cite{mubarak2022overview}, and Tashkeela \cite{zerrouki2017tashkeela} focus on language modeling or specific tasks rather than full-page OCR evaluation. Handwriting datasets including HistoryAr \cite{pantke2014historical}, IFN/ENIT \cite{pechwitz2002ifn}, KHATT \cite{mahmoud2014khatt}, APTI \cite{slimane2009new}, and Muharaf \cite{saeed2024muharaf} emphasize word/line recognition over document structure analysis.

\begin{table}[t]
\centering
\small
\begin{tabular}{@{}p{4cm}r@{}}
\toprule
\textbf{Domain} & \textbf{Total Samples} \\
\midrule
PDF to Markdown & 33 \\
Layout & 2,100 \\
Line Detection & 378 \\
Line Recognition & 378 \\
Table Recognition & 456 \\
Image to Text & 3,760 \\
Charts to DataFrame & 576 \\
Diagram to Json & 226 \\
VQA & 902 \\
\midrule
\textbf{Total} & \textbf{8,809} \\
\bottomrule
\end{tabular}
\caption{Distribution of samples across different domains in our dataset. A more detailed count for different sub-domains and data sources is in Appendix \ref{sec:appendix1}.}
\label{tab:dataset-distribution}
\end{table}

\noindent Arabic table recognition faces challenges from merged cells and RTL formatting \cite{pantke2014historical}. While methods like GTE \cite{zheng2021global}, GFTE \cite{li2021gfte}, CascadeTabNet \cite{prasad2020cascadetabnet}, TableNet \cite{paliwal2019tablenet}, and TableFormer \cite{nassar2022tableformer} advance Latin table detection, their effectiveness on Arabic documents remains unproven. Document conversion pipelines (CCS \cite{staar2018corpus}, Tesseract \cite{smith2007overviewtesseract}, Docling \cite{auer2024docling}, Surya \cite{paruchuri2024surya}, Marker \cite{paruchuri2024marker}, MinerU \cite{wang2024mineru}, PaddleOCR \cite{du2020paddleocr}) lack Arabic-specific optimizations for segmentation and diacritic handling \cite{mahmoud2018online, badam2024benjamin}. This highlights the critical need for comprehensive Arabic OCR benchmarks addressing text recognition, table detection, and layout parsing.

\section{KITAB-Bench}
\begin{figure*}[htbp]
    \centering
    \small
    \includegraphics[width=0.95\textwidth]{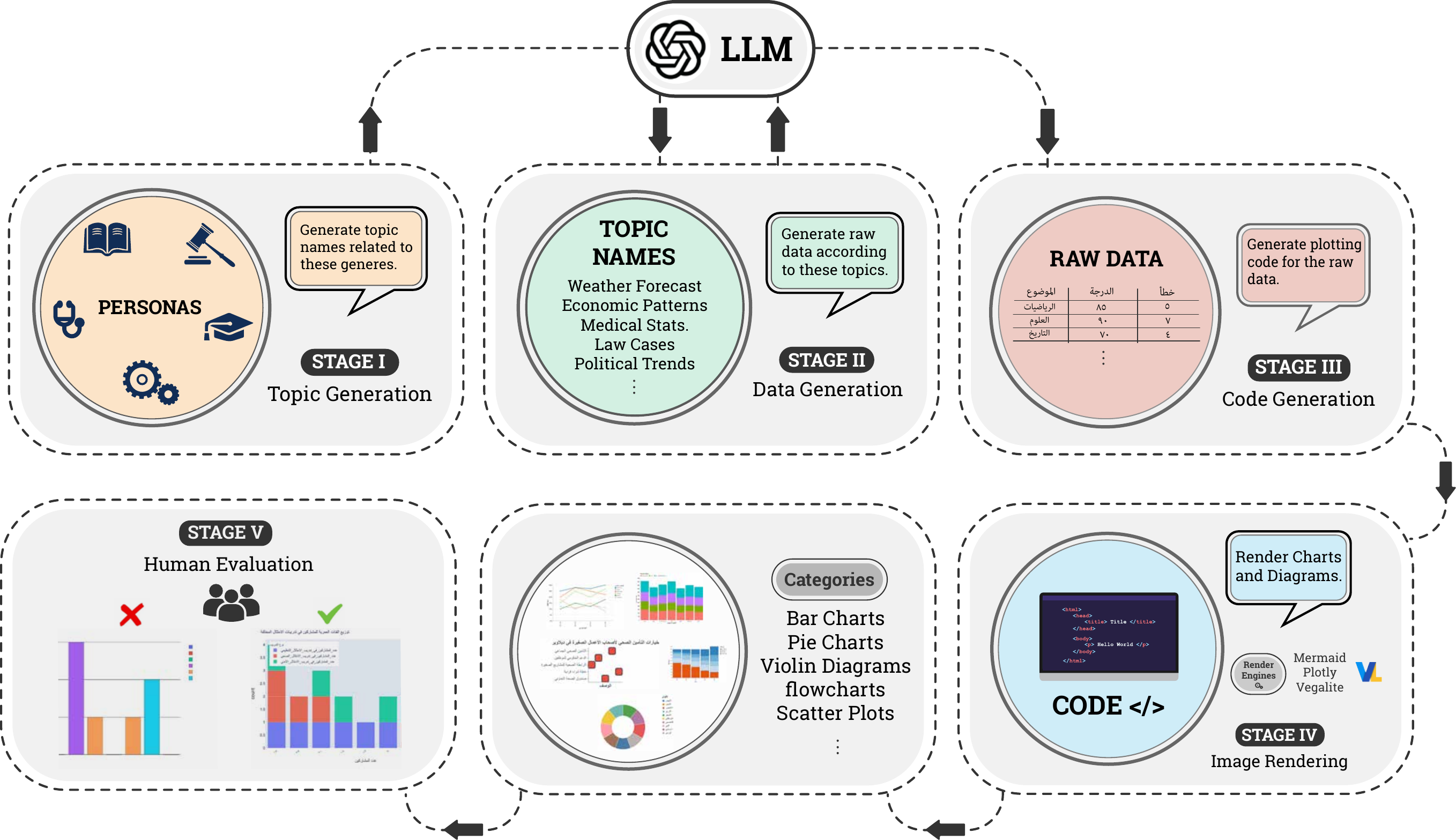}
    \caption{Synthetic Data Generation Pipeline: A 5-stage process using LLMs to generate topics, create raw data, produce visualization code, render charts, and perform human evaluation for quality control.}
    \label{fig:data_creation_pipeline}
\end{figure*}

Our methodology offers a novel approach to benchmarking Arabic OCR systems via a comprehensive data collection strategy and a systematic evaluation framework. We gather curated samples from existing Arabic document datasets, manually collected and annotated PDFs, and employ a five-phase LLM-assisted human-in-the-loop pipeline (Figure~\ref{fig:data_creation_pipeline}) to generate diverse supplementary content. Our evaluation framework spans nine specialized tasks, enabling thorough assessment of OCR performance across various document processing challenges and providing a robust benchmark for Arabic document understanding tasks.

\subsection{PDF Data Collection}

We curated 33 diverse PDFs from online sources in academia, medicine, law, and literature. To ensure challenging cases, we selected documents featuring richly formatted tables with extensive color usage, merged cells, Arabic numerals, historical texts, watermarks, and handwritten annotations. Each PDF averaged three pages, and we then manually annotated them. This dataset comprehensively captures real-world complexities, making it a valuable benchmark for PDF-to-Markdown conversion.



\subsection{LLM-Assisted Data Generation Pipeline}
\label{subsection:data_generation}
To generate data for charts, diagrams and tables, we implemented a five-phase LLM-assisted generation pipeline with human validation at critical stages, as illustrated in Figure~\ref{fig:data_creation_pipeline}. 
\textit{In Phase I (Topic Generation)}, our system employs an LLM to generate diverse topic names across multiple domains. This phase incorporates various personas (academic, legal, medical, technical) to ensure broad coverage of document types. 
\textit{Phase II (Data Generation)} transforms the validated topics into structured raw data. The LLM generates content following Arabic linguistic and formatting conventions across various domains. 
\textit{In Phase III (Code Generation)}, the system converts the validated raw data into plotting code, with special attention to Arabic text rendering requirements and RTL content management. 
\textit{Phase IV (Image Rendering)} utilizes specialized rendering engines (Mermaid, Plotly, Vegalite, HTML) to create visual representations while maintaining Arabic text integrity.

\textit{The final phase (Human Evaluation)} implements rigorous quality control through expert validation. Evaluators filter charts, tables and diagrams based on detected anomalies and ensure adherence to Arabic-specific document conventions. This phase is crucial for maintaining the high quality of our benchmark dataset.
\setlength{\parindent}{0pt}

\subsection{Dataset Statistics}

Our benchmark dataset comprises over 8,809 samples across 9 major domains and 36 sub-domains, representing a comprehensive collection of Arabic document types for OCR evaluation. As detailed in Table \ref{tab:dataset_distribution}, the dataset combines carefully curated samples from established datasets, manually annotation PDFs, and synthetically generated content created through our LLM-assisted pipeline (Figure \ref{fig:data_creation_pipeline}). The Image-to-Text portion (3,760 samples) includes data from historical documents (HistoryAr \cite{pantke2014historical}), handwritten text collections (Khatt \cite{mahmoud2014khatt}, ADAB \cite{adab2019bench}, Muharaf \cite{saeed2024muharaf}), and scene text (EvAREST \cite{hassan2021arabicEvArEST}), while layout detection comprises 2,100 samples from BCE-Arabic-v1 \cite{saad2016bce} and DocLayNet \cite{doclaynet2022}.

For layout analysis, we incorporated 1,700 samples from BCE-Arabic-v1 dataset \cite{saad2016bce}, 400 samples from DocLayNet dataset \cite{doclaynet2022} focusing on financial, academic, legal, and patent documents. The line detection and recognition tasks contains 378 samples each from self-developed dataset. We further enriched the dataset with 500 samples from PATS-A01 \cite{muhtaseb2010pats} benchmark to ensure diverse representation.
\begin{table}[htbp]
\centering
\renewcommand{\arraystretch}{0.9}
\setlength{\tabcolsep}{3.2pt}
\resizebox{\linewidth}{!}{
\begin{tabular}{l l c c c}
\toprule
\textbf{Task} & \textbf{Metric} & \textbf{Surya} & \textbf{Tesseract} & \textbf{EasyOCR} \\
\midrule
\multirow{2}{*}{Detection} & mAP@50 & \textbf{79.67} & 46.39 & 68.02 \\
& mAP@0.5:0.95 & 27.40 & 14.30 & \textbf{32.74} \\
\midrule
\multirow{2}{*}{Recognition} & WER & 1.01 & 1.00 & \textbf{0.53} \\
& CER & 0.87 & 0.66 & \textbf{0.20} \\
\bottomrule
\end{tabular}
}
\caption{Performance of different models on Line Detection and Line Recognition Task on our Benchmark}
\label{tab:line-detection}
\end{table}
For handwritten text recognition, we assembled a comprehensive collection of 1,000 samples combining datasets from Khatt \cite{mahmoud2014khatt} (both paragraph and line-level annotations), Adab \cite{adab2019bench}, Muharaf \cite{saeed2024muharaf}, and OnlineKhatt \cite{mahmoud2018online}. The benchmark also includes specialized content from ISI-PPT \cite{wu2017iccv} (500 samples), and Hindawi \cite{elfilali2023hindawi} (200 samples) for various document types. Scene text understanding is supported by 800 samples from EvArest \cite{hassan2021arabicEvArEST}, providing real-world context diversity. A detailed table showing all the dataset is provided in the Appendix \ref{sec:appendix1}.

A significant portion of our dataset consists of synthetically generated content, including 576 samples for Charts-to-DataFrame (spanning 16 different chart types), 422 samples for Diagram-to-Code (covering sequence diagrams, flowcharts, and tree maps), 456 samples for Tables-to-CSV/HTML, and 902 samples for VQA tasks. These synthetic samples were generated through our five-phase LLM-assisted human-in-the-loop pipeline (Figure \ref{fig:data_creation_pipeline}). Every sample in our dataset - whether from existing sources or newly generated - underwent validation by native Arabic speakers before inclusion in the final benchmark. This rigorous validation, reinforced by expert review and automated checks, ensures high quality and authenticity across all domains. A detailed analysis is in Appendix \ref{sec:appendix3}.

\section{Experiments}
Our experimental evaluation comprehensively assesses the capabilities of current OCR systems and state-of-the-art vision-language models (VLMs) across different Arabic and multilingual document understanding tasks. Figure~\ref{fig:pipeline_task_overview} illustrates the nine distinct tasks in our evaluation framework.

We evaluate three categories of systems: VLMs, traditional OCR systems, and specialized document processing tools. For VLMs, we include both closed-source models like \texttt{gpt-4o-2024-08-06}, \texttt{gpt-4o-mini-2024-07-18} \cite{hurst2024gpt, achiam2023gpt}, and \texttt{gemini-2.0-flash} \cite{team2024gemini, google2025gemini}, as well as open-source alternatives such as \texttt{Qwen2-VL-7B}~\cite{wang2024qwen2}, \texttt{Qwen2.5-VL-7B} \cite{Qwen2.5-VL}, and the \texttt{AIN-7B} \cite{heakl2025ain}. Traditional OCR approaches in our evaluation include \texttt{Surya} \cite{paruchuri2024surya}, \texttt{Tesseract} \cite{smith2007overviewtesseract}, \texttt{EasyOCR} \cite{easyocr}, and \texttt{PaddleOCR} \cite{li2022paddleocr, du2021paddleocr}. For specialized document processing tasks, we employ systems like \texttt{Docling} \cite{auer2024docling}, and \texttt{Marker} \cite{paruchuri2024marker}. Layout detection capabilities are evaluated using methods implemented in \texttt{Surya-layout} \cite{paruchuri2024surya}, \texttt{Yolo-doclaynet} \cite{zhao2024doclayout} from \texttt{MinerU} \cite{ wang2024mineru}, and \texttt{RT-DETR} \cite{zhao2023detrs} based method in Docling \cite{auer2024docling}.

\begin{table}[b]
\centering
\renewcommand{\arraystretch}{1.1}
\setlength{\tabcolsep}{2.5pt}
\footnotesize
\small
\begin{tabular}{l l | c c c}
\toprule
\textbf{Dataset} & \textbf{Metric} & \textbf{Surya} & \textbf{Yolo-doc-} & \textbf{Detr} \\
& & & \textbf{laynet} & \textbf{(docling)} \\
\midrule
\multirow{5}{*}{BCE} & mAP@0.5 & 0.506 & 0.470 & \textbf{0.750} \\
& mAP@0.5:0.95 & 0.381 & 0.369 & \textbf{0.566} \\
& Precision & \textbf{0.751} & 0.608 & 0.626 \\
& Recall & 0.593 & 0.592 & \textbf{0.725} \\
& F1 Score & 0.635 & 0.585 & \textbf{0.654} \\
\midrule
\multirow{5}{*}{DocLayNet} & mAP@0.5 & 0.675 & 0.404 & \textbf{0.758} \\
& mAP@0.5:0.95 & 0.469 & 0.335 & \textbf{0.541} \\
& Precision & \textbf{0.782} & 0.527 & 0.635 \\
& Recall & 0.856 & 0.503 & \textbf{0.770} \\
& F1 Score & 0.799 & 0.499 & \textbf{0.670} \\
\bottomrule
\end{tabular}
\caption{Performance comparison of layout detection models using different evaluation metrics}
\label{tab:layout-detection}
\end{table}
\subsection{Evaluation Frameworks and Metrics}
Our evaluation framework comprises nine specialized tasks designed to assess different aspects of Arabic OCR systems, as demonstrated in Figure~\ref{fig:pipeline_task_overview}. Each task addresses specific challenges in Arabic document processing. For this reason, we employ task-specific metrics to evaluate different aspects of document understanding.

\textbf{PDF-to-Markdown:} It evaluates the conversion of Arabic PDFs to structured markdown while preserving the text and table structure. Since both table and text structure are important, for evaluating PDF to Markdown conversion quality, we propose MARS (Markdown Recognition Score), which combines chrF \cite{popovic2015chrf} with Tree-Edit-Distance-based Similarity (TEDS) \cite{zhong2020image} :
\begin{equation}
\label{eq:mars}
    \text{MARS} = \alpha \cdot \text{chrF}_3 + (1-\alpha) \cdot \text{TEDS}(T_a, T_b)
\end{equation}
where $\alpha$ ($0 \leq \alpha \leq 1$) is the weight. $T_a$ represent predicted table structure  and $T_b$ the ground truth structure.

\textbf{Table Recognition:} We evaluate table extraction using both HTML and CSV formats, where HTML format (evaluated using TEDS \cite{zhong2020image}) preserves rich structural information including cell spans and hierarchical relationships crucial for complex Arabic tables, while CSV format (evaluated using Jaccard Index \ref{eq:jaccard}) focuses on raw data extraction optimized for machine processing and data analysis pipelines. This dual-format evaluation ensures systems can both maintain complex table structures for human readability and provide clean, structured data for automated processing, specifically important for RAG based systems.
\begin{equation}
    \label{eq:jaccard}
    J(P,G) = \frac{|P \cap G|}{|P \cup G|} = \frac{|P \cap G|}{|P| + |G| - |P \cap G|}
\end{equation}
where $|P \cap G|$ represents the number of exact matching cells between predicted and ground truth tables, and $|P \cup G|$ represents the total number of unique cells across both tables.

\textbf{Chart-to-Dataframe:} This task evaluates extracting structured data from Arabic charts into machine-readable dataframes. Systems must accurately parse numerical values, text labels, and preserve data relationships across chart types (bar, line, pie). We use the Structuring Chart-oriented Representation Metric (SCRM) \cite{xia2024chartx}—which combines type recognition, topic understanding, and structural numerical fidelity (see Appendix \ref{sec:appendix4})—and also propose our own CharTeX (Chart Extraction Score) metric. CharTeX combines the chrF scores for chart type and topic with the jaccord index for the dataframe, using fuzzy matching (80\% threshold) when columns do not exactly align.
\vspace{-3pt}
\begin{equation} \label{eq:codm} 
\text{Metric} = \alpha J_{type} + \beta J_{topic} + (1-\alpha-\beta)J_{data} \end{equation}

Here, $J_{type}$ and $J_{topic}$ denote the chrF scores between the predicted and ground-truth chart type and topic, while $J_{data}$ measures the structural similarity of the predicted and ground-truth JSON data.

\textbf{Diagram-to-JSON: } This task evaluates the conversion of Arabic flowcharts and technical diagrams into JSON while preserving semantic relationships and technical specifications. We propose CODM (Code-Oriented Diagram Metric), extending SCRM~\cite{xia2024chartx}, with the same fomulation as in Eq~\ref{eq:codm}. More detail about this metric is provided in Appendix \ref{sec:appendix4}.

\begin{table*}[t]
\centering
\setlength{\tabcolsep}{4pt}
\small
\begin{tabular}{l l c c c c c}
\toprule
& & \multicolumn{2}{c}{\textbf{Table Extraction}} & \multicolumn{3}{c}{\textbf{End-to-End PDF}} \\
\cmidrule(lr){3-4} \cmidrule(lr){5-7}
\textbf{Model Group} & \textbf{Models} & \textbf{TEDS (HTML)} & \textbf{Jaccard (CSV)} & \textbf{CHrF (Text)} & \textbf{TEDS (Table)} & \textbf{MARS} \\
\midrule
\multirow{3}{*}{\makecell[l]{Closed}} 
& GPT-4o & \textbf{85.76} & \textbf{66.36} & 69.62 & \textbf{60.61} & 65.12 \\
& GPT-4o-mini & 69.32 & 49.50 & 56.59 & 52.69 & 54.64 \\
& Gemini-2.0-Flash & 83.08 & 65.55 & \textbf{75.75} & 55.55 & \textbf{65.65} \\
\midrule
\multirow{3}{*}{\makecell[l]{Open}}
& Qwen2-VL-7B & 57.83 & 40.20 & 40.30 & 2.54 & 21.42 \\
& Qwen2.5-VL-7B & 59.31 & 59.58 & 69.21 & 11.65 & 40.43 \\
& AIN-7B & 75.94 & 64.83 & 56.52 & 49.32 & 52.92 \\
\midrule
\multirow{4}{*}{\makecell[l]{Framework}}
& Tesseract & \makecell{28.23$^D$\\38.64$^I$} & \makecell{14.85$^D$\\16.04$^I$} & 59.91$^D$ & 45.44$^D$ & 52.68$^D$ \\[3pt]
& EasyOCR & \makecell{49.10$^D$\\39.09$^I$} & \makecell{23.83$^D$\\17.88$^I$} & 57.46$^D$ & 51.12$^D$ & 54.29$^D$ \\[3pt]
& Surya & 50.15$^M$ & 70.42$^M$ & 58.38$^M$ & 44.29$^M$ & 51.34$^M$ \\
\bottomrule
\multicolumn{7}{l}{\footnotesize $^D$Docling \cite{auer2024docling} pipeline \quad $^I$Img2Table \cite{img2table} pipeline \quad $^M$Marker \cite{paruchuri2024marker} pipeline}
\end{tabular}
\caption{Performance comparison of different models for table extraction and end-to-end PDF to markdown conversion tasks on our benchmark.}
\label{tab:table-pdf-tasks}
\end{table*}
\textbf{Image-to-Text: } This task assess the basic text recognition capabilities across different Arabic fonts and styles, including the handling of cursive script connections, diacritical marks, and various text orientations. We use we use Character Error Rate (CER) and Word Error Rate (WER). For a predicted text sequence $\hat{y}$ and ground truth sequence $y$, CER is computed as:
$
\label{eq:cer}
\text{CER} = \frac{\text{L}(y, \hat{y})}{|y|}
$,
where $\text{L}(y, \hat{y})$ is the Levenshtein distance between character sequences and $|y|$ is the ground truth length. WER is calculated the same way with words as the unit of error.

\textbf{Visual Question Answering: } Tests the ability of models to understand and reason about Arabic document content, we evaluate using standard accuracy for MCQ questions and exact word match. 

\textbf{Line Detection: } Focuses on the accurate identification and processing of individual text lines in Arabic documents. We evaluate using mean Average Precision (mAP) at different Intersection over Union (IoU) thresholds: mAP@0.5 and mAP@0.5:0.95, which assess the localization accuracy of detected text lines.


\textbf{Layout Detection: } Assesses document structure analysis capabilities, including the identification of headers, paragraphs, and complex layout elements in Arabic documents. Performance is measured using mAP@0.5 and mAP@0.5:0.95 for localization accuracy, complemented by Precision, Recall, and F1 scores to evaluate the overall detection quality across different layout components.

All metrics are computed on our diverse benchmark dataset, which encompasses various document types and complexity levels in both Arabic and multilingual contexts. Table~\ref{tab:doc-understanding-metric-model} provides a detailed mapping of tasks, metrics, and evaluated systems.
\begin{table}[htbp]
\centering
\setlength{\tabcolsep}{5pt}
\resizebox{\linewidth}{!}{
\begin{tabular}{l l c c c}
\toprule
\textbf{Group} & \textbf{Models} & \textbf{CHrF $\uparrow$} & \textbf{CER $\downarrow$} & \textbf{WER $\downarrow$} \\
\midrule
\multirow{3}{*}{\makecell[l]{Closed}} 
& GPT-4o & 61.01 & 0.31 & 0.55 \\
& GPT-4o-mini & 47.21 & 0.43 & 0.71 \\
& Gemini-2.0-Flash & 77.95 & \textbf{0.13} & 0.32 \\
& Azure & 50.97 & 0.52 & 0.69 \\
\midrule
\multirow{3}{*}{\makecell[l]{Open}}
& Qwen2VL-7B & 33.94 & 1.48 & 1.55 \\
& Qwen2.5VL-7B & 49.23 & 1.20 & 1.41 \\
& AIN-7B & \textbf{78.33} & 0.20 & \textbf{0.28} \\
& Qaari & 39.77	& 1.80 & 1.93 \\
& Gemma3 & 30.02 & 1.05 & 1.45 \\
& ArabicNagout & 30.52 & 4.37 & 4.67 \\
\midrule
\multirow{4}{*}{\makecell[l]{Framework}}
& Tesseract & 39.62 & 0.54 & 0.84 \\
& EasyOCR & 45.47 & 0.58 & 0.89 \\
& Paddle & 16.73 & 0.79 & 1.02 \\
& Surya & 20.61 & 4.95 & 5.61 \\
\bottomrule
\end{tabular}
}
\caption{Performance comparison of models for OCR (image to text) tasks on our benchmark. A detailed performance comparison among different open-source dataset is available in Appendix \ref{sec:appendix2}}
\label{tab:ocr-tasks}
\end{table}

\subsection{Experimental Setup}
We implement our evaluation pipeline with careful consideration of hyperparameters for different metrics. All experiments use NVIDIA A100 GPUs. For VLMs, we use their official implementations or API endpoints. Traditional OCR systems are evaluated using pre-trained models provided by the frameworks. For PDF-to-Markdown evaluation metric MARS \ref{eq:mars}, we choose $\alpha=0.5$ and $\alpha =0.5$ and $\beta = 0.2$ for Diagram-to-JSON evaluation metric CODM. We average the results over multiple runs, with performance comparisons shown in different tables (Table \ref{tab:line-detection}, \ref{tab:layout-detection}, \ref{tab:table-pdf-tasks}, \ref{tab:ocr-tasks}, and \ref{tab:visual-tasks}).

\begin{table*}[htbp]
\centering
\renewcommand{\arraystretch}{1.2}
\setlength{\tabcolsep}{5pt}
\footnotesize
\resizebox{\textwidth}{!}{
\begin{tabular}{l l c c c c c c c c}
\toprule
\multirow{2}{*}{\textbf{Group}} & \multirow{2}{*}{\textbf{Model}} & \multicolumn{2}{c}{\textbf{Chart}} & \textbf{Diagram} & \multicolumn{5}{c}{\textbf{Visual QA}} \\
\cmidrule(lr){3-4} \cmidrule(lr){5-5} \cmidrule(lr){6-10}
& & SCRM & CharTeX & CODM & MTVQA$^O$ & ChartsVQA$^M$ & DiagramsVQA$^M$ & PATDVQA$^M$ & Average \\
\midrule
\multirow{3}{*}{Closed} 
& GPT-4o & 68.6 & 45.95 & 61.6 & 32.00 & 77.00 & 85.29 & 82.50 & 69.19\\
& GPT-4o-mini & 67.2 & 43.33 & 61.4 & 26.80 & 58.00 & 83.33 & 80.00 & 62.03\\
& Gemini-2.0-Flash & \textbf{71.4} & \textbf{56.28} & \textbf{71.8} & \textbf{35.00} & 72.00 & \textbf{88.24} & 75.50 & 67.68 \\
\midrule
\multirow{3}{*}{Open} 
& Qwen2-VL-7B & 56.6 & 21.59 & 63.0 & 19.60 & 59.00 & 82.35 & 77.50 & 59.61\\
& Qwen2.5-VL-7B & 36.2 & 22.08 & 59.2 & 23.00 & 74.00 & 79.41 & 74.50 & 62.72 \\
& AIN-7B & 66.6 & 34.61 & 66.40 & 31.50 & 75.00 & 85.29 & \textbf{87.00} & \textbf{69.69}\\
\bottomrule
\end{tabular}
}
\caption{Model Performance on Chart Understanding, Diagram Parsing, and Visual Question Answering Tasks. For VQA tasks, $O$ denotes open-ended question type from MTVQA \cite{tang2024mtvqa} dataset and $M$ denotes MCQ type questions.}
\label{tab:visual-tasks}
\end{table*}
\section{Results and Discussion}
In this section, we present a comprehensive evaluation of different models across different tasks of our framework. The results provide a clear distinction between the performance of closed-source models, open-source models, and framework-based solutions, revealing both their strengths and limitations. We observe very clear performance gap between closed and open-source solutions. While closed-source models like Gemini-2.0-Flash consistently outperform other models almost all the tasks.

\subsection{Charts, Diagrams, and VQA}
Table [\ref{tab:visual-tasks}] presents model performance across different chart and diagram understanding tasks, evaluated using SCRM and CharTeX (for charts), and VQA-based accuracy metrics. Among closed-source models, Gemini-2.0 achieves the highest performance on chart understanding metrics, scoring 71.4\% on SCRM and 56.28\% on CharTeX. The performance gap between Gemini-2.0 and GPT-4o is particularly pronounced in CharTeX evaluation (10.33\%) compared to SCRM (2.8\%). Open-source models shows a significant limitation in complex chart understanding. While their SCRM scores remain competitive, both Qwen variants score below 23\% on CharTeX evaluation.
The visual question-answering results reveal an important exception to the general closed-source advantage. AIN achieves 87\% on PATDVQA, surpassing Gemini-2.0 by 11.5\%. AIN also shows competitive performance on MTVQA (31.50\%), which is similar to GPT-4o and ~4\% better than GPT-4o-mini. This shows that open-source models can be competitive with closed-source alternatives.

\subsection{Layout and Lines: Document Structure}
Our evaluation of document structure understanding reveals distinct performance patterns across layout detection and line processing tasks. In layout detection (Table~\ref{tab:layout-detection}), RT-DETR \cite{zhao2023detrs} achieves superior overall performance with mAP@0.5 scores of 0.750 and 0.758 on BCE (arabic only) and DocLayNet (english) datset respectively. However, Surya \cite{paruchuri2024surya} demonstrates higher precision (0.782 on DocLayNet, 0.751 on BCE), despite lower recall rates. This trade-off suggests that different architectures optimize for different aspects of layout detection.

The line processing results (Table~\ref{tab:line-detection}) highlight a clear contrast between detection and recognition capabilities. While Surya excels in detection with a mAP@0.50 of 79.67\%, EasyOCR demonstrates superior recognition performance (WER: 0.53, CER: 0.20). This inverse relationship between detection and recognition performance across models indicates a fundamental challenge in optimizing both capabilities simultaneously. Notably, Tesseract shows consistent but lower performance across both metrics, suggesting that newer architectures have made significant improvements over traditional approaches.
We also observe that no single model excels at both detection and recognition, which requires for hybrid solutions.

\subsection{Tables, OCR, and PDF-to-Markdown}

Across table extraction tasks (Table~\ref{tab:table-pdf-tasks}), closed-source models maintain a clear advantage, with GPT-4o achieving 85.76\% TEDS and 66.36\% Jaccard scores. Among open-source models, AIN (75.94\% TEDS) significantly outperforms Qwen variants, while specialized frameworks like Surya achieve competitive results (70.42\% Jaccard) through targeted pipelines. 

For OCR tasks, we evaluated GPT-4o ~\cite{hurst2024gpt}, Gemini-2.0-Flash ~\cite{google2025gemini}, Azure OCR~\cite{microsoft_azure_ocr} in closed model; Qaari~\cite{QariOCRCollection2025}, Gemma3~\cite{team2025gemma}, ArabicNagout~\cite{rashad2024arabic} and AIN~\cite{heakl2025ain} in open source models and Tesseract~\cite{smith2007overviewtesseract}, EasyOCR~\cite{easyocr}, PaddleOCR~\cite{li2022paddleocr} and SuryaOCR~\cite{paruchuri2024surya} in frameworks (Table~\ref{tab:ocr-tasks}).
Gemini-2.0-Flash leads with the lowest error rates (CER: 0.13, WER: 0.32). Notably, AIN matches this performance level (WER: 0.28), while traditional OCR frameworks like EasyOCR and Tesseract show moderate performance (CER: 0.58, 0.54). The significant performance drop in Paddle (CER: 0.79) and Surya (CER: 4.95) highlights the challenges in developing robust OCR systems.

End-to-end document processing (Table \ref{tab:table-pdf-tasks}) reveals the largest gaps between approaches. Closed-source models maintain consistent performance (GPT-4o: 65.12\% MARS, Gemini-2.0: 65.65\% MARS), while open-source models show substantial degradation (Qwen2-VL-7B: 21.42\% MARS). Framework approaches achieve better stability, with Tesseract and EasyOCR scoring above 50\% MARS, suggesting that specialized pipelines can partially bridge the gap with larger models in complete document processing tasks.

\begin{figure}[h]
    \centering
    \includegraphics[width=\linewidth]{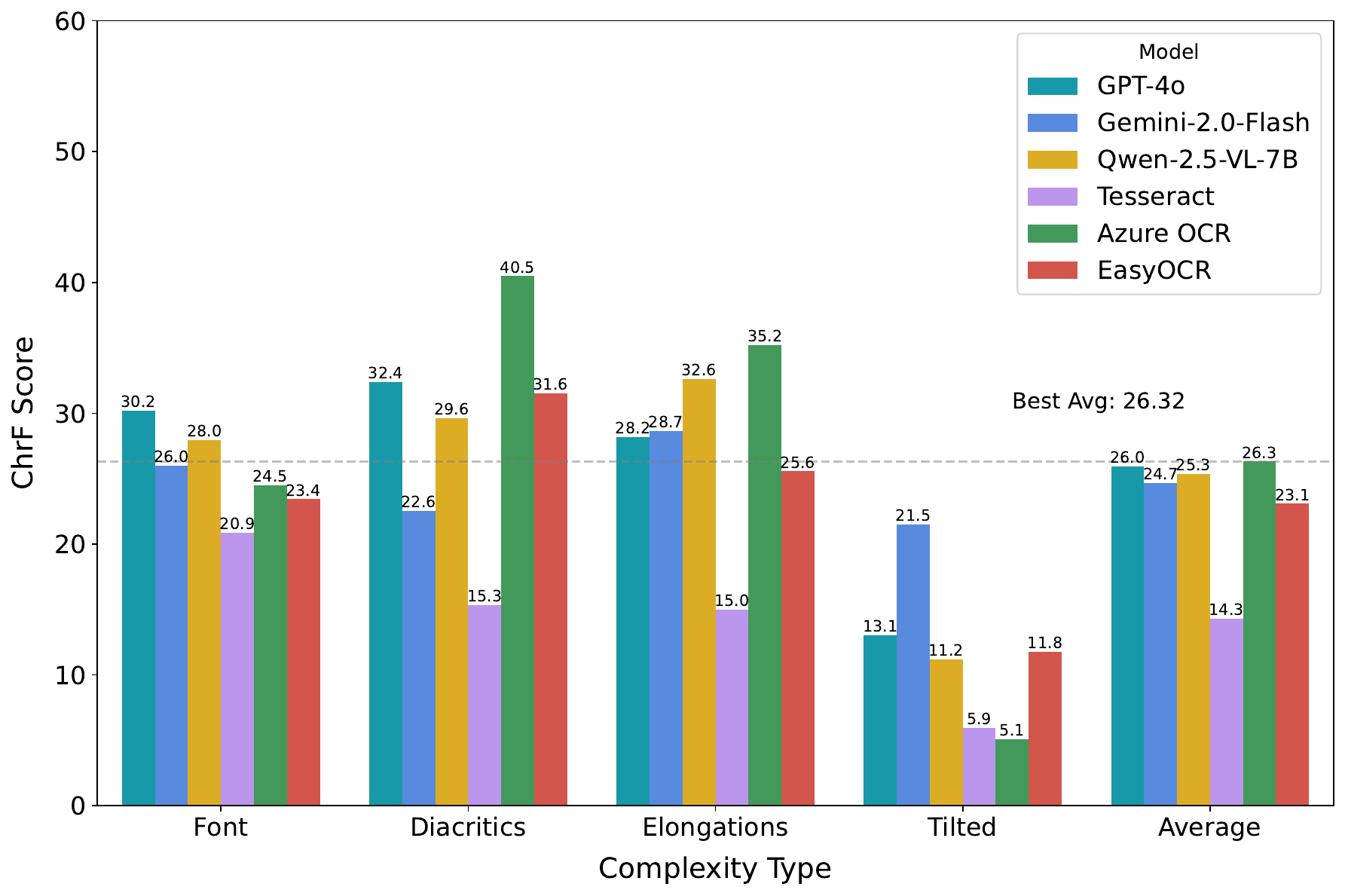}
    \caption{ChrF by model on Arabic text variations}
    \label{fig:models-compare}
\end{figure}

Our comprehensive evaluation demonstrates that while closed-source models maintain superior performance over open-source models across most tasks, specialized frameworks like Surya, RT-DETR Layout, and EasyOCR achieve competitive performance in targeted scenarios like table extraction, layout detection, and text recognition respectively. However, this framework advantage significantly diminishes in end-to-end pdf-to-markdown tasks where the integration capabilities of large models prove crucial, as evidenced by the performance gaps between commercial VLMs and traditional systems like EasyOCR, Surya and Tesseract in End-to-End PDF task (Table \ref{tab:table-pdf-tasks}).

\subsection{Performance on Challenging Cases}

To evaluate model performance across different complexities of Arabic texts, we manually selected 104 samples representing four challenging categories: font variations, diacritics, text elongations, and tilted text. The ChrF score comparison (Figure~\ref{fig:models-compare}) reveals distinct performance patterns across models, with GPT-4o demonstrating superior font handling (30.2) and leading in challenging tilted text recognition (13.1), while Azure OCR excels remarkably in diacritics recognition (40.5) and text elongations (35.2), indicating specialized Arabic script optimizations. The overall performance analysis shows GPT-4o leading at 26.0 average ChrF score, followed closely by Azure (26.3), Qwen2.5-VL-7B (25.3), and Gemini-2.0-Flash (24.7), while traditional OCR systems struggle significantly with Tesseract particularly challenged by diacritics (15.3) and tilted text (5.9). This analysis reveals that no single model excels across all Arabic text complexities, with specialized systems like Azure demonstrating domain-specific strengths in diacritics and elongation handling, while modern VLMs show more consistent performance but struggle with orientation variations, underscoring the need for Arabic-specific optimizations and highlighting the substantial performance gap between modern VLMs and traditional OCR approaches.

\subsection{Model Performance across Chart Types}

\begin{figure}[h]
    \centering
    \includegraphics[width=\linewidth]{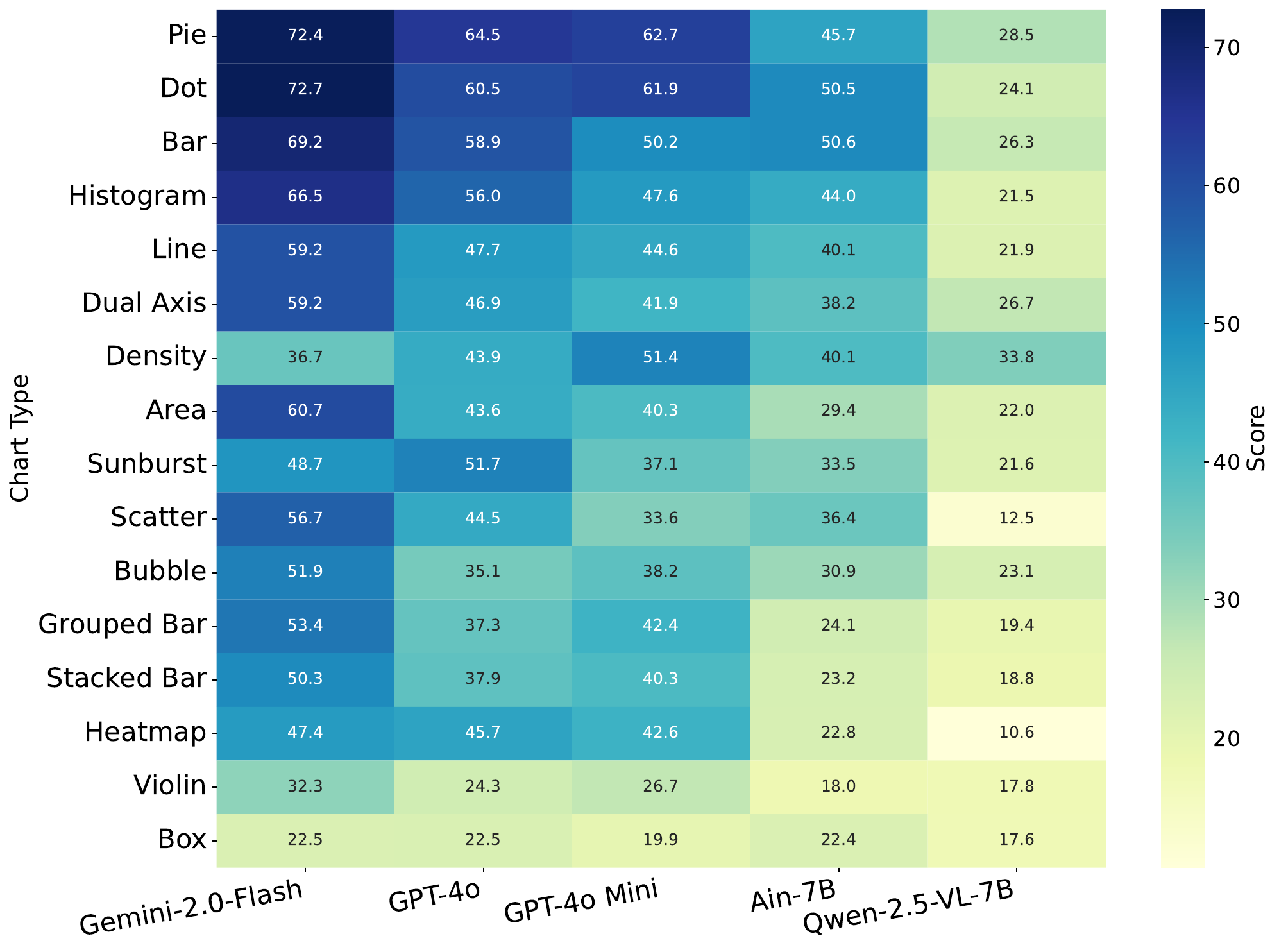}
    \caption{ChartEx results across different charts type.}
    \label{fig:charts-compare}
\end{figure}

The CharTeX evaluation across 16 different chart types reveals significant performance variations based on chart complexity and structural characteristics (Figure~\ref{fig:charts-compare}). Gemini-2.0-Flash demonstrates superior performance across most chart types, particularly excelling in simple geometric charts like Pie (72.4), Dot (72.7), and Bar Charts (69.2), while complex statistical visualizations like Violin Plots (32.3) and Box Plots (22.5) present significant challenges for all models. Simple chart types with clear boundaries consistently achieve higher scores across all models, with grouped and stacked bar charts showing intermediate performance levels around 40-50, indicating that while structural complexity affects extraction accuracy, the familiarity of bar chart formats provides some resilience. This pattern suggests that Arabic chart understanding faces particular difficulties with charts requiring statistical interpretation and continuous data representation, highlighting that current models perform best on charts with discrete, clearly separated data elements rather than continuous or overlapping visual representations.

\section{Conclusion}

We introduce a comprehensive benchmark for Arabic OCR that fills the gap in standardized evaluation frameworks for Arabic document processing. Our dataset of 8,809 samples across nine major domains is the most diverse collection assembled for OCR evaluation, incorporating handwritten, scanned, synthetic, and scene text, as well as complex tables, charts, and end-to-end pdf-to-markdown. This framework extends beyond simple text recognition to include structural document analysis and enables systematic assessment of OCR performance across various fonts, styles, and layouts.

\section{Limitations and Future Directions}

Despite its strengths, KITAB-Bench lacks coverage of low-resource dialects and institutional scans such as historical, governmental, and financial records. Future work should address OCR limitations in structural fidelity for tables and charts through richer datasets, refined metrics, and cross-lingual deep learning methods to enable robust and generalizable Arabic multimodal OCR. Moreover, current models often fail to generalize across domains and layouts, emphasizing the need for adaptable architectures and domain-specific fine-tuning.


\bibliography{acl_natbib}

\appendix

\section{Source of the Existing Dataset Collection}
\label{sec:appendix1}

Our benchmark integrates diverse data sources to ensure comprehensive coverage of Arabic document types. As detailed in Table \ref{tab:dataset-distribution}, the dataset combines manually curated samples, synthetic data generated through our LLM-assisted pipeline (Figure \ref{fig:data_creation_pipeline}), and existing publicly available datasets. Key sources include:

\begin{itemize}
\item {Handwritten Text: KHATT (paragraph and line-level annotations), ADAB, Muharaf, and OnlineKhatt.}

\item {Historical Documents: HistoryAr and HistoricalBooks.}

\item {Scene Text: EvAREST for real-world context diversity.}

\item {Layout Analysis: BCE-Arabic-v1 and DocLayNet.}

\item {Synthetic Content: 576 chart samples (16 types) and 422 diagram samples generated via our five-phase pipeline (Section 3.2).}
\end{itemize}
The dataset emphasizes domain diversity, covering academic, medical, legal, financial, and technical documents. All samples underwent rigorous validation by native Arabic speakers to ensure linguistic and structural accuracy.

\section{Detailed Performance Comparison}
\label{sec:appendix2}

Table \ref{tab:VLM_COMP} provides granular performance metrics for VLMs and OCR frameworks across 12 Arabic text recognition datasets. Gemini-2.0-Flash demonstrates exceptional robustness on synthetic datasets (CER: 0.01 on PATS), while AIN-7B excels in historical manuscript recognition (CER: 0.26 on HistoryAr). Traditional OCR systems like Tesseract show limitations in handwritten text (CER: 1.26 on HistoryAr), highlighting the need for script-specific optimizations.

\section{Data Analysis}
\label{sec:appendix3}

\begin{figure}[p]
    \centering
    \includegraphics[width=1.0\textwidth]{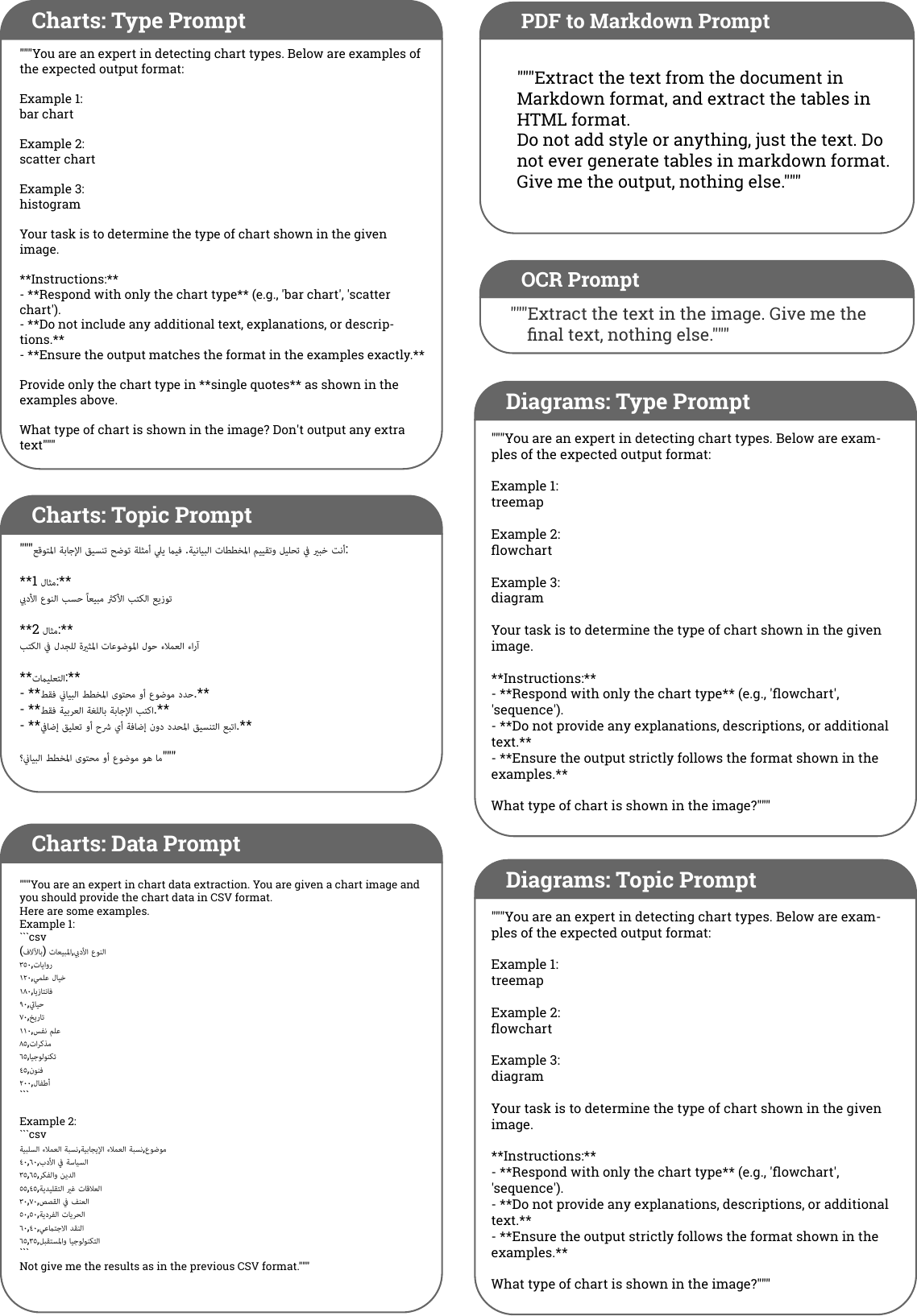}
    \caption{Prompts for Different Task Categories.}
    \label{fig:prompta}
\end{figure}

\begin{figure}[htbp]
    \centering
    \includegraphics[width=1.0\textwidth]{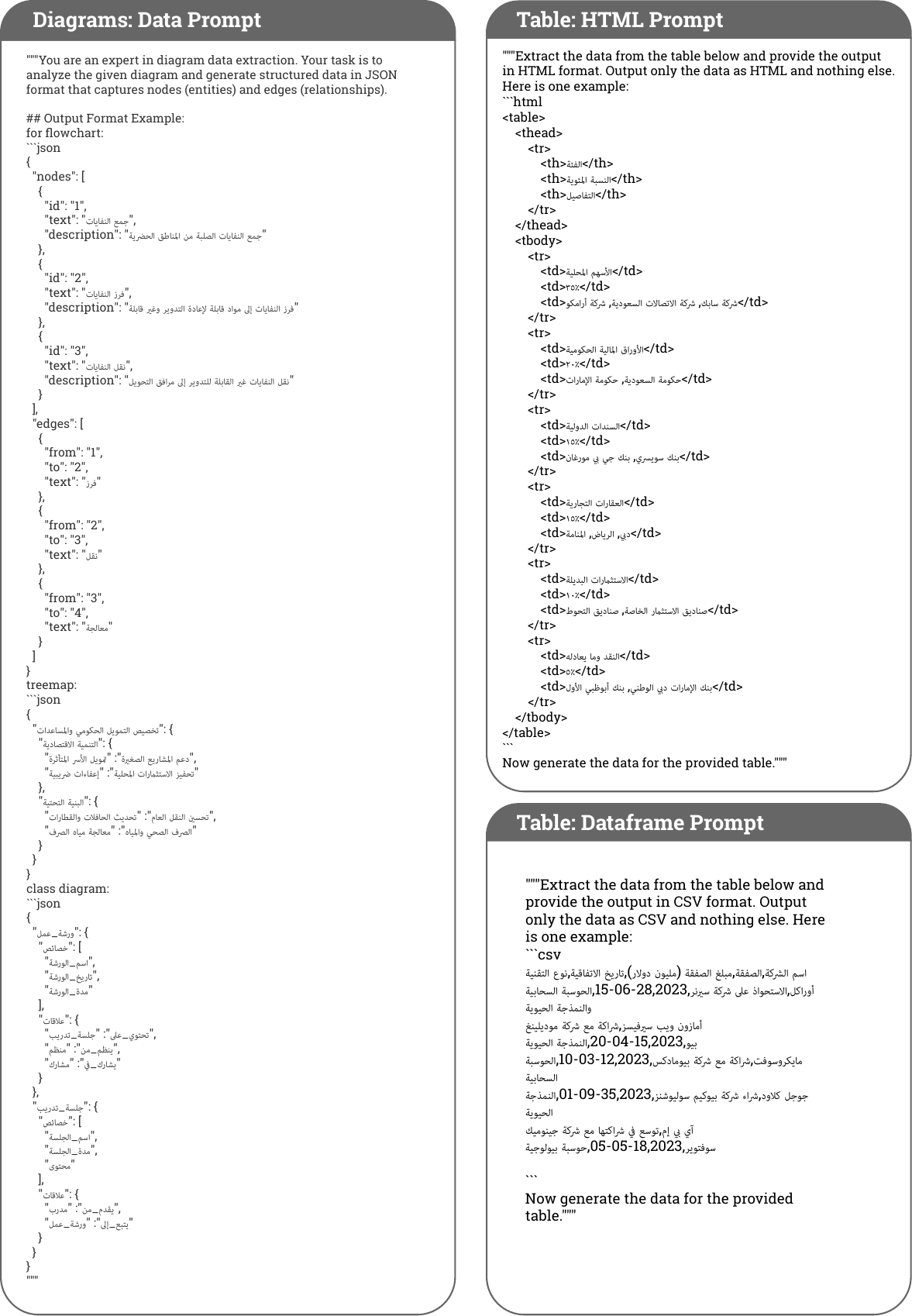}
    \caption{Prompts for Diagrams and Tables.}
    \label{fig:promptb}
\end{figure}

Our data generation pipeline (Figure \ref{fig:data_creation_pipeline}) produced 1,502 high-quality synthetic samples - comprising 576 graphs, 422 diagrams, and 456 tables, through LLM-assisted generation guided by domain-specific instructions (Figures \ref{fig:prompta} and \ref{fig:promptb}) that ensured alignment with Arabic linguistic norms. During the human validation phase, 18\% of initial outputs were discarded due to issues like right-to-left formatting errors and semantic inconsistencies. The resulting dataset offers diverse and balanced coverage, featuring 21 Arabic calligraphic styles, 36 sub-domains spanning financial reports to technical manuals, and complex structures such as merged cells in 43\% of tables and dual-axis configurations in 29\% of charts.

\section{Evaluation Metrics}
\label{sec:appendix5}

\subsection{Tasks Models and Metrics}
\label{sec:appendix4}

Table \ref{tab:doc-understanding-metric-model} maps evaluation tasks to corresponding models and metrics. The framework evaluates nine core capabilities:

\begin{itemize}
    \item Structural Understanding: Layout detection (mAP), line detection (IoU)
    \item Content Extraction: Text recognition (CER), table parsing (TEDS)
    \item Semantic Reasoning: VQA accuracy, chart-to-dataframe conversion (SCRM)
    \item Specialized metrics like MARS (
    $\alpha$=0.5) address the dual requirements of text fidelity and structural preservation in PDF-to-Markdown conversion.
\end{itemize}

\subsection{Structuring Chart-oriented Representation Metric (SCRM)}

The Structuring Chart-oriented Representation Metric (SCRM) evaluates chart understanding through three weighted components:

\begin{equation}
    \text{SCRM} = 0.4J_{\text{type}} + 0.3J_{\text{topic}} + 0.3J_{\text{data}}
\end{equation}

where $J_{\text{type}}$ measures chart type recognition accuracy using Edit Distance, $J_{\text{topic}}$ evaluates chart topic identification using Edit Distance, and $J_{\text{data}}$ measures Mean Relative Error with and Error Thrsholding criteria.

For entity comparison in $J_{\text{type}}$ and $J_{\text{topic}}$, we use the chrF character-based metric which captures partial matches effectively. For data comparison, value similarity is computed using relative error:
\[
e(p, q) = \frac{|\text{Value}_{\text{pred}}^p - \text{Value}_{\text{GT}}^q|}{\text{Value}_{\text{GT}}^q}
\]

\subsection{Chart Extraction Score (CharTeX)}

To evaluate chart data extraction quality, we propose CharTeX (Chart Extraction Score), which combines character-level text similarity with structural data assessment:

\begin{equation}
    \text{CharTeX} = \alpha J_{\text{type}} + \beta J_{\text{topic}} + (1 - \alpha - \beta) J_{\text{data}}
\end{equation}

Where $\alpha = 0.05$ and $\beta = 0.10$ in our implementation, reflecting the relative importance of each component where $J_{\text{type}}$ evaluates chart type recognition using chrF score (5\% weight), $J_{\text{topic}}$ assesses topic identification using chrF score (10\% weight), and $J_{\text{data}}$: measures structural data extraction accuracy using fuzzy matching (85\% weight).

CharTeX improves upon SCRM by introducing structure-aware fuzzy matching (95\% threshold) and leveraging the Hungarian algorithm for optimal alignment. In contrast to SCRM's reliance on \textit{(entity, value)} triplet matching, CharTeX incorporates column-level semantics and chrF-based scoring, enhancing robustness to text variations and structural discrepancies, particularly critical for Arabic charts with complex layouts. This design mitigates SCRM’s sensitivity to superficial mismatches and its disregard for tabular structure.

\subsection{Markdown Recognition Score (MARS)}

To evaluate the quality of PDF-to-Markdown conversion, we propose the Markdown Recognition Score (MARS), defined as:
\[
\text{MARS} = \alpha \cdot \text{chrF3} + (1 - \alpha) \cdot \text{TEDS}(T_a, T_b)
\]
where $\alpha \in [0, 1]$ is set to 0.5 to balance text fidelity and structural accuracy. Here, $T_a$ and $T_b$ denote the predicted and ground truth table structures, respectively.

MARS jointly captures character-level accuracy using chrF3, ideal for OCR tasks requiring fine-grained text recognition, and hierarchical layout preservation via TEDS, which quantifies the tree-edit distance between table structures. By assigning equal weight to both components, MARS offers a robust metric that reflects both semantic and structural fidelity in document conversion. As both chrF3 and TEDS are established in prior work, MARS inherits their theoretical validity without the need for further empirical justification.

\subsection{Code-Oriented Diagram Metric (CODM)}

The Code-Oriented Diagram Metric (CODM) extends SCRM with a graph-theoretic foundation specifically designed for diagrams where structural relationships are paramount:

\begin{equation}
    \text{CODM} = 0.5J_{\text{topology}} + 0.2J_{\text{topic}} + 0.3J_{\text{semantics}}
\end{equation}

Where $J_{\text{topology}}$ evaluates diagram type (50\%) using edit distance, $J_{\text{topic}}$ assesses topic identification (20\%) using edit distance, and $J_{\text{semantics}}$ measures diagram structure through Graph Edit Distance (30\%).

This metric converts both predicted and ground truth diagram data into graph structures, where nodes represent entities and edges represent relationships. This approach effectively evaluates both node-edge relationships and semantic labels in technical diagrams such as flowcharts, class diagrams, and sequence diagrams.

Further, domain-specific prompts are used to guide model responses for accurate metric calculation. For instance, sequence diagrams require strict adherence to Arabic UML notation standards during evaluation, ensuring fair assessment across different diagram conventions.

\begin{table*}[htbp]
\centering
\small
\begin{tabular}{@{}p{2.5cm}p{2.5cm}p{5.0cm}rrr@{}}
\hline
\textbf{Domain} & \textbf{Sub-Domain} & \textbf{Dataset Source} & \textbf{Original} & \textbf{Selected} & \textbf{Total} \\
\hline
PDF to Markdown & General & Manual & 33 & 33 & 33 \\
\hline
Layout Detection & Docs & BCE-Arabic-v1 \cite{saad2016bce} & 1.9k & 1,700 & \multirow{2}{*}{2,100} \\
 & & DocLayNet \cite{doclaynet2022} & 80k & 400 & \\
\hline
Line Detection & Docs & Manual & 375 & 378 & 378 \\
\hline
Line Recognition & Docs & Manual & 375 & 378 & 378 \\
\hline
Table Recognition & Financial & Pixmo \cite{deitke2024molmo} & 490 & 456 & 456 \\
\hline
\multirow{13}{*}{Image to Text} & \multirow{2}{*}{Synthetic} & PATS \cite{muhtaseb2010pats} & 21.6k & 500 & \multirow{13}{*}{3,760} \\
 & & SythenAR & 39.1k & 500 & \\
 & \multirow{2}{*}{Historical} & HistoryAr  \cite{pantke2014historical} & 1.5k & 200 & \\
 & & HistoricalBooks & 40 & 10 & \\
 & Hand. Paragraph & Khatt \cite{mahmoud2014khatt} & 2.72k & 200 & \\
 & Hand. Word & ADAB \cite{adab2019bench} & 15k & 200 & \\
 & \multirow{3}{*}{Hand. Line} & Muharaf \cite{saeed2024muharaf} & 24.5k & 200 & \\
 & & OnlineKhatt \cite{mahmoud2018online} & 8.5k & 200 & \\
 & & Khatt \cite{mahmoud2014khatt} & 13.4k & 200 & \\
 & PPT & ISI-PPT \cite{wu2017iccv} & 86.5k & 500 & \\
 & \multirow{2}{*}{Blogs} & ArabicOCR & 20.3k & 50 & \\
 & & Hindawi \cite{elfilali2023hindawi} & 79k & 200 & \\
 & Scene & EvAREST \cite{hassan2021arabicEvArEST} & 5.59k & 800 & \\
\hline
\multirow{16}{*}{Charts to DataFrame} & Bar & Synthetic & 100 & 61 & \multirow{16}{*}{576} \\
 & Line & Synthetic & 100 & 43 & \\
 & Pie & Synthetic & 100 & 56 & \\
 & Box & Synthetic & 100 & 31 & \\
 & Violin & Synthetic & 100 & 36 & \\
 & Area & Synthetic & 50 & 29 & \\
 & SunBurst & Synthetic & 30 & 15 & \\
 & Dot & Synthetic & 30 & 15 & \\
 & Dual Axis & Synthetic & 20 & 26 & \\
 & Density Curve & Synthetic & 10 & 5 & \\
 & Bubble & Synthetic & 20 & 13 & \\
 & Grouped Bar & Synthetic & 50 & 60 & \\
 & Stacked Bar & Synthetic & 50 & 82 & \\
 & Histogram & Synthetic & 100 & 70 & \\
 & HeatMap & Synthetic & 10 & 11 & \\
 & Scatter & Synthetic & 100 & 23 & \\
\hline
\multirow{7}{*}{Diagram to Json} & Sequence & Synthetic & 50 & 46 & \multirow{7}{*}{226} \\
 & Funnel & Synthetic & 20 & 52 & \\
 & Class & Synthetic & 20 & 30 & \\
 & Network & Synthetic & 20 & 18 & \\
 & Venn & Synthetic & 20 & 7 & \\
 & FlowChart & Synthetic & 100 & 112 & \\
 & TreeMap & Synthetic & 100 & 157 & \\
\hline
\multirow{4}{*}{VQA} & Diagrams & Manual & 102 & 102 & \multirow{4}{*}{902} \\
 & Charts & Manual & 105 & 100 & \\
 & News Letter & PATD \cite{bouressace2019printed} & 2.42k & 200 & \\
 & Scene & MTVQA & 818 & 500 & \\
\hline
\multicolumn{3}{l}{\textbf{Total Dataset Size}} & -- & \multicolumn{2}{r}{8,809} \\
\hline
\end{tabular}
\caption{Dataset Distribution Across Different Domains, sub-domains and Data Source}
\label{tab:dataset_distribution}
\end{table*}

\begin{table*}[htbp]
\centering
\label{tab:results-vlm}
\renewcommand{\arraystretch}{1.2} 
\setlength{\tabcolsep}{6pt} 

\footnotesize

\begin{tabular}{l r | cc cc cc cc | cc}
\toprule
\multirow{2}{*}{\textbf{Dataset}} & \multirow{2}{*}{\textbf{Size}} 
& \multicolumn{2}{c}{\textbf{GPT-4o}} 
& \multicolumn{2}{c}{\textbf{GPT-4o-mini}} 
& \multicolumn{2}{c}{\textbf{Azure OCR}} 
& \multicolumn{2}{c|}{\textbf{Gemini-2.0-Flash}}
& \multicolumn{2}{c}{\textbf{Qwen2-VL}} \\
& & CER & WER & CER & WER & CER & WER & CER & WER & CER & WER \\
\midrule
PATS & 500 & 0.23 & 0.30 & 0.53 & 0.71 & 0.03 & 0.10 & 0.01 & 0.02 & 1.02 & 1.02 \\
SythenAR & 500 & 0.09 & 0.20 & 0.14 & 0.32 & 0.10 & 0.27 & 0.07 & 0.17 & 0.59 & 1.13 \\
HistoryAr & 200 & 0.51 & 0.82 & 0.67 & 0.96 & 0.24	& 0.68 & 0.28 & 0.64 & 3.46 & 2.86 \\
HistoricalBooks & 10 & 0.41 & 0.76 & 0.59 & 0.88 & 0.29	& 0.71 & 0.05 & 0.22 & 1.90 & 2.16 \\
Khatt & 200 & 0.45 & 0.74 & 0.64 & 0.91 & 0.83 & 0.92 & 0.19 & 0.45 & 1.12 & 5.04 \\
Adab & 200 & 0.30 & 0.73 & 0.35 & 0.83 & 0.99 & 0.99 & 0.19 & 0.56 & 0.63 & 1.08 \\
Muharaf & 200 & 0.56 & 0.90 & 0.63 & 0.94 & 0.52 & 0.82 & 0.33 & 0.69 & 3.57 & 2.87 \\
OnlineKhatt & 200 & 0.29 & 0.63 & 0.41 & 0.76 & 0.72 & 0.85 & 0.17 & 0.44 & 1.30 & 2.01 \\
ISI-PPT & 500 & 0.08 & 0.18 & 0.15 & 0.31 & 0.98 & 0.98 & 0.06 & 0.15 & 1.03 & 1.06 \\
ArabicOCR & 50 & 0.06 & 0.26 & 0.16 & 0.46 & 0.01 & 0.11 & 0.00 & 0.02 & 1.25 & 1.50 \\
Hindawi & 200 & 0.34 & 0.56 & 0.48 & 0.71 & 0.06 & 0.28 & 0.01 & 0.04 & 1.82 & 2.05 \\
EvArest & 800 & 0.20 & 0.38 & 0.25 & 0.51  & 0.32 & 0.50 & 0.18 & 0.36 & 0.41 & 0.95 \\
\midrule
  & 3,760 & {0.31} & {0.55} & 0.43 & 0.71 & 0.52 & 0.69 & 0.13 & 0.32 & 1.48 & 1.20 \\
\bottomrule
\end{tabular}

\vspace{4mm} 

\resizebox{\linewidth}{!}{
\begin{tabular}{l r | cc cc cc | cc cc cc}
\toprule
\multirow{2}{*}{\textbf{Dataset}} & \multirow{2}{*}{\textbf{Size}} 
& \multicolumn{2}{c}{\textbf{Qwen2.5-VL}} & \multicolumn{2}{c}{\textbf{AIN}} & \multicolumn{2}{c|}{\textbf{Qari}} 
& \multicolumn{2}{c}{\textbf{Tesseract}} & \multicolumn{2}{c}{\textbf{Surya}} & \multicolumn{2}{c}{\textbf{Paddle}} \\
& & CER & WER & CER & WER & CER & WER & CER & WER & CER & WER & CER & WER \\
\midrule
PATS & 500 & 0.98 & 1.03 & 0.26 & 0.36 & 0.00 & 0.00 & 0.14 & 0.28 & 4.66 & 4.67 & 0.77 & 1.00 \\
SythenAR & 500 & 1.68 & 1.69 & 0.21 & 0.40 & 0.04 & 0.16 & 0.31 & 0.72 & 4.82 & 7.90 & 0.80 & 1.01 \\
HistoryAr & 200 & 3.48 & 3.39 & 0.47 & 0.83 & 0.26 & 0.54 & 0.72 & 1.26 & 10.32 & 12.78 & 0.79 & 1.01 \\
HistoricalBooks & 10 & 0.67 & 0.97 & 0.33 & 0.72 & 0.84 & 0.88 & 0.74 & 0.99 & 6.81 & 6.30 & 0.71 & 1.00 \\
Khatt & 200 & 1.60 & 1.80 & 0.07 & 0.22 & 0.61 & 1.12 & 0.67 & 1.06 & 4.25 & 3.77 & 0.76 & 1.00 \\
Adab & 200 & 0.91 & 1.11 & 0.00 & 0.01 & 1.00 & 1.00 & 1.00 & 1.14 & 7.28 & 8.71 & 0.88 & 1.15 \\
Muharaf & 200 & 2.40 & 2.74 & 0.61 & 0.96 & 0.38 & 0.54 & 0.77 & 1.22 & 6.19 & 7.48 & 0.80 & 1.01 \\
OnlineKhatt & 200 & 1.52 & 1.53 & 0.36 & 0.70 & 0.03 & 0.12 & 0.59 & 1.20 & 6.71 & 6.95 & 0.78 & 1.03 \\
ISI-PPT & 500 & 1.27 & 1.39 & 0.36 & 0.54 & 0.52 & 0.53 & 0.31 & 0.64 & 4.25 & 3.77 & 0.81 & 1.03 \\
ArabicOCR & 50 & 0.02 & 0.08 & 1.00 & 1.00 & 0.01 & 0.01 & 0.01 & 0.01 & 2.75 & 3.58 & 0.77 & 1.00 \\
Hindawi & 200 & 0.27 & 0.42 & 1.00 & 1.00 & 0.11 & 0.15 & 0.31 & 0.72 & 0.15 & 0.20 & 0.76 & 1.00 \\
EvArest & 800 & 4.65 & 4.75 & 0.19 & 0.36 & 0.30 & 0.32 & 0.85 & 1.02 & 5.91 & 3.86 & 0.89 & 1.04 \\
\midrule
  & 3,760 & 1.80 & 1.93 & 0.28 & 0.54 & {0.20} & {0.58} & 0.89 & 0.79 & 4.95 & 5.61 & 0.79 & 1.02 \\
\bottomrule
\end{tabular}}
\caption{Performance comparison of Large Vision-Language Models on KITAB-Bench (lower is better).}
\label{tab:VLM_COMP}
\end{table*}

\begin{table*}[htbp]
\centering
\small
\resizebox{\textwidth}{!}{
\begin{tabular}{p{2.5cm}p{4cm}p{2.5cm}p{2.5cm}p{4cm}}
\toprule
\textbf{Task} & \textbf{Metrics} & \textbf{Open LLMs} & \textbf{Closed LLMs} & \textbf{OCR Systems} \\
\midrule
\multicolumn{5}{l}{\textit{Document Understanding Tasks}} \\
\midrule
PDF to Markdown & chrF + TEDS & -- & -- & \begin{tabular}[t]{@{}l@{}} Docling \\ Marker \\ MinerU \\ PDF-Extract-Kit \end{tabular} \\
\midrule
Layout Detection & \begin{tabular}[t]{@{}l@{}} mAP@0.5 \\ mAP@0.5:0.95 \\ Precision \\ Recall \\ F1 \end{tabular} & -- & -- & \begin{tabular}[t]{@{}l@{}} Surya \\ Yolo-doclaynet (MinerU) \\ Detr (docling) \end{tabular} \\
\midrule
Line Detection & \begin{tabular}[t]{@{}l@{}} mAP@0.5 \\ mAP@0.5:0.95 \end{tabular} & -- & -- & \begin{tabular}[t]{@{}l@{}} Surya \\ Tesseract \\ EasyOCR \end{tabular} \\
\midrule
Line Recognition & WER, CER & -- & -- & \begin{tabular}[t]{@{}l@{}} Surya \\ Tesseract \\ EasyOCR \end{tabular} \\
\midrule
\multicolumn{5}{l}{\textit{Table Understanding Tasks}} \\
\midrule
Tables Recognition (HTML) & TEDS \cite{zhong2019image} & \begin{tabular}[t]{@{}l@{}} Qwen2-VL \\ Qwen2.5-VL \\ AIN \\ PaliGemma \end{tabular} & \begin{tabular}[t]{@{}l@{}} GPT-4o \\ GPT-4o-mini \\ Gemini-2.0-Flash \end{tabular} & \begin{tabular}[t]{@{}l@{}} Docling[EasyOCR] \\ Docling[Tesseract] \\ Marker \\ Img2Table[EasyOCR] \\ Img2Table[Tesseract] \end{tabular} \\
\midrule
Tables Recognition (CSV) & Jaccard Index & \begin{tabular}[t]{@{}l@{}} Qwen2-VL \\ Qwen2.5-VL \\ AIN \\ PaliGemma \end{tabular} & \begin{tabular}[t]{@{}l@{}} GPT-4o \\ GPT-4o-mini \\ Gemini-2.0-Flash \end{tabular} & \begin{tabular}[t]{@{}l@{}} Docling[EasyOCR] \\ Docling[Tesseract] \\ Marker \\ Img2Table[EasyOCR] \\ Img2Table[Tesseract] \end{tabular} \\
\midrule
\multicolumn{5}{l}{\textit{Visual Understanding Tasks}} \\
\midrule
Image to Text & \begin{tabular}[t]{@{}l@{}} CER, WER \\ chrF, BLEU \\ METEOR \end{tabular} & \begin{tabular}[t]{@{}l@{}} Qwen2-VL \\ Qwen2.5-VL \\ AIN-7B \\ PaliGemma \end{tabular} & \begin{tabular}[t]{@{}l@{}} GPT-4o \\ GPT-4o-mini \\ Gemini-2.0-Flash \end{tabular} & \begin{tabular}[t]{@{}l@{}} Docling[EasyOCR] \\ Docling[Tesseract] \\ Marker \\ Img2Table[EasyOCR] \\ Img2Table[Tesseract] \end{tabular} \\
\midrule
Charts to DataFrame & SCRM \cite{xia2024chartx, xia2023structchart} & \begin{tabular}[t]{@{}l@{}} Qwen2-VL \\ Qwen2.5-VL \\ AIN \\ PaliGemma \end{tabular} & \begin{tabular}[t]{@{}l@{}} GPT-4o \\ GPT-4o-mini \\ Gemini-2.0-Flash \end{tabular} & -- \\
\midrule
Diagram to Json & SCRM & \begin{tabular}[t]{@{}l@{}} Qwen2-VL \\ Qwen2.5-VL \\ AIN-7B \\ PaliGemma \end{tabular} & \begin{tabular}[t]{@{}l@{}} GPT-4o \\ GPT-4o-mini \\ Gemini-2.0-Flash \end{tabular} & -- \\
\midrule
VQA & \begin{tabular}[t]{@{}l@{}} Accuracy + \\ Word Match Score \end{tabular} & \begin{tabular}[t]{@{}l@{}} Qwen2-VL \\ Qwen2.5-VL \\ AIN-7b \\ PaliGemma \end{tabular} & \begin{tabular}[t]{@{}l@{}} GPT-4o \\ GPT-4o-mini \\ Gemini-2.0-Flash \end{tabular} & -- \\
\bottomrule
\end{tabular}
}
\caption{Comprehensive evaluation metrics and models for document understanding tasks. The table is organized into three main categories: document understanding, table understanding, and visual understanding tasks. Each task is evaluated using specific metrics and implemented across various models and OCR systems.}
\label{tab:doc-understanding-metric-model}
\end{table*}

\clearpage

\clearpage

\end{document}